\documentclass[sigconf]{acmart}

\setcopyright{none}
\renewcommand\footnotetextcopyrightpermission[1]{} 
\usepackage{amsmath,amsfonts,bm}

\def\eqref#1{equation~\ref{#1}}

\def\1{\bm{1}}

\DeclareMathAlphabet{\mathsfit}{\encodingdefault}{\sfdefault}{m}{sl}
\SetMathAlphabet{\mathsfit}{bold}{\encodingdefault}{\sfdefault}{bx}{n}

\usepackage{hyperref}
\usepackage{url}
\usepackage{graphicx}
\usepackage{amsmath}
\usepackage{bbm}
\usepackage{multirow}

\newcommand{\mb}{\mathbf}
\newcommand{\bs}{\boldsymbol}
\newcommand{\mc}{\mathcal}

\newtheorem{definition}{\textsc{Definition}}

\newcommand{\gcn}{\textsc{Gcn}}
\newcommand{\our}{\textsc{Iso-CapsNet}}
\newcommand{\isonn}{\textsc{IsoNN}}
\newcommand{\fastisonn}{\textsc{FastIsoNN}}

\newcommand{\freq}{\textsc{Freq}}
\newcommand{\wl}{\textsc{WL}}
\newcommand{\cnn}{\textsc{Cnn}}
\newcommand{\gin}{\textsc{Gin}}
\newcommand{\autoencoder}{\textsc{AE}}
\newcommand{\sdbn}{\textsc{Sdbn}}
\newcommand{\capsule}{\textsc{CapsNet}}
\newcommand{\dgcnn}{\textsc{DGCnn}}
\newcommand{\diffpool}{\textsc{DiffPool}}

\begin{document}

\title{{\our}: Isomorphic Capsule Network for Brain Graph Representation Learning}

\author{Jiawei Zhang}
\email{jiawei@ifmlab.org}
\affiliation{%
  \institution{IFM Lab\\ Department of Computer Science\\ University of California, Davis}
  \city{Davis}
  \state{California}
  \country{USA}
  \postcode{95616}
}

\begin{abstract}

Brain graph representation learning serves as the fundamental technique for brain diseases diagnosis. Great efforts from both the academic and industrial communities have been devoted to brain graph representation learning in recent years. The isomorphic neural network ({\isonn}) introduced recently can automatically learn the existence of sub-graph patterns in brain graphs, which is also the state-of-the-art brain graph representation learning method by this context so far. However, {\isonn} fails to capture the orientations of sub-graph patterns, which may render the learned representations to be useless for many cases. In this paper, we propose a new {\our} (Isomorphic Capsule Net) model by introducing the graph isomorphic capsules for effective brain graph representation learning. Based on the capsule dynamic routing, besides the sub-graph pattern existence confidence scores, {\our} can also learn other sub-graph rich properties, including \textit{position}, \textit{size} and \textit{orientation}, for calculating the class-wise digit capsules. We have compared {\our} with both classic and state-of-the-art brain graph representation approaches with extensive experiments on four brain graph benchmark datasets. The experimental results also demonstrate the effectiveness of {\our}, which can outperform the baseline methods with significant improvements.

\end{abstract}


\settopmatter{printfolios=true}
\maketitle

\section{Introduction}\label{sec:introduction}


\begin{figure}[t]
    \centering
    	\includegraphics[width=0.9\linewidth]{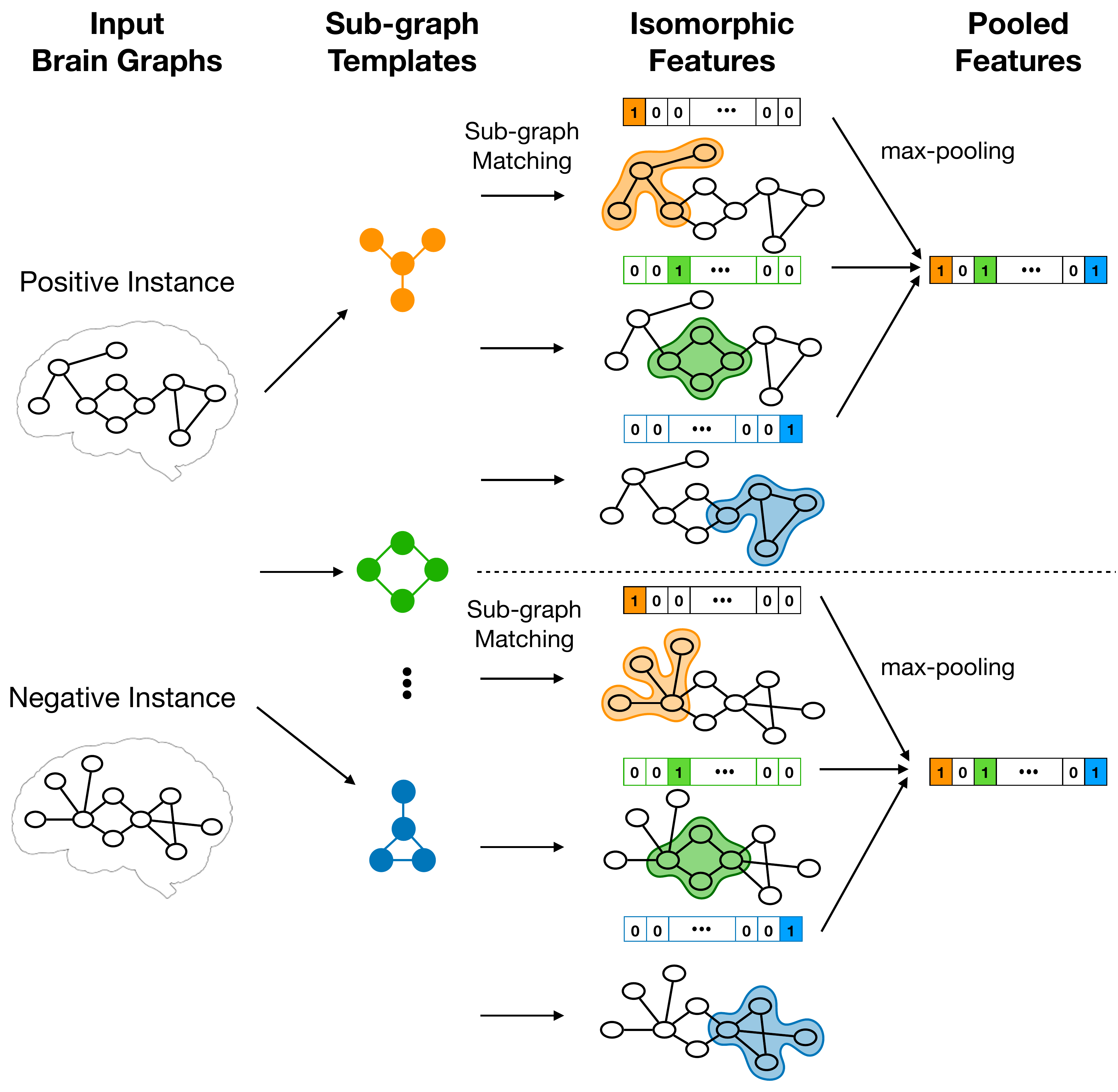}
    	\caption{An Illustration about the weakness of the {\isonn} model in learning brain graph representations. Given two different brain graph instances (with different labels), without considering the sub-graph pattern orientations, {\isonn} will learn identical isomorphic representations for these two input brain graphs, since the sub-graph patterns (in different orientations) exist at the same areas of the input graph instances. It will render {\isonn} fail to work.}\label{fig:example}
\end{figure}

Brain graph denotes the graph structured data illustrating the activity correlations among different areas in human brains \cite{Kong_Brain_14}. Human brain graphs usually come from the brain raw imaging data, scanned with EEG, CT or fMRI \cite{Mitra_Analysis_99}. Via necessary data pre-processing, de-noising, filtering, normalization, partition and statistical analysis, human brain raw imaging data can be transformed and reduced into brain graph structured representations \cite{Stam_Graph_08,Lee_Identifying_17}. Great efforts have been denoted to brain graph analysis in recent years \cite{Fallani_Graph_14}, which helps illustrate the mysterious working mechanism of human brains. Meanwhile, brain graph analysis can also help the early diagnosis of many brain diseases, e.g., Alzheimer's disease, Parkinson's Disease, Schizophrenia and Depression \cite{Bassett_Human_09,Menon_Large_11,Cao_Tensor_14}.

In recent years, brain graph representation learning has also become an important research topic \cite{Wang_Structural_17,Lee_Deep_20,Meng_Isomorphic_NIPS_19}, which aims at learning the lower-dimensional feature representations of the human brain data. For instance, \cite{Wang_Structural_17} proposes to apply the CNN model to the brain graph inputs, prior to that necessary node re-ordering is necessary; \cite{Lee_Deep_20} introduces a brain graph representation learning model by extending the graph convolutional network. Different from other research areas, interpretability of the brain graph learning results is very important. \cite{Meng_Isomorphic_NIPS_19} proposes the {\isonn} (isomorphic neural net) by extending the classic sub-graph mining techniques \cite{Yan_GSpan_02} to brain graph representation learning. Meanwhile, different from conventional sub-graph mining, {\isonn} learns sub-graph patterns from brain graphs automatically, and the representation features denote the existence confidence scores of sub-graphs in the input brain graph. Such learned sub-graph patterns will also serve as the interpretation of the learning results.

The brain graph isomorphic feature representations learned by {\isonn} can identify the existence of sub-graph patterns in the input graph data effectively. However, such learned representations can hardly capture the orientations or poses of the sub-graph patterns actually, which may render the model to be defective for many cases. To help illustrate the problem clearly, we also provide an example in Figure~\ref{fig:example}. As shown in the figure, we feed the model with two different input brain graphs with different labels, where \textit{positive} denotes ``\textit{healthy} patient'' and \textit{negative} denotes ``\textit{sick} patient'', respectively. Via the sub-graph templates learned in {\isonn} (regardless of the different sub-graph orientations), both of these two input graph data instances can achieve identical graph isomorphic feature representations, which will render the model fail to classify the input data instances.

To solve such a problem, in this paper, we will introduce a new brain graph representation learning model, namely {\our} (Isomorphic Capsule Net). Via sub-graph pattern learning and matching, {\our} learns the graph isomorphic capsules as the representations, which captures not only the existence of the sub-graph patterns but also other extra rich properties (such as the pose, size and orientation) of the sub-graphs in the input brain graph instances. To be more specific, based on the learned graph isomorphic features, in capsule layers, we will use the overall length of the instantiation parameter vectors to denote the existence of subgraphs, and force the orientation of the vectors to represent the aforementioned properties of these subgraphs. Such capsule layers will be initialized with graph isomorphic representations and iteratively updated with the dynamic routing algorithm \cite{NIPS2017_6975}.

Here, we summarize the contributions of this paper as follows:
\begin{itemize}

\item In this paper, we introduce a new brain graph representation learning model, which integrates the advantages of both {\isonn} and capsule network together. The learned representations can indicate both the existence and the aforementioned basic properties of sub-graph patterns in the input brain graphs, which are denoted by the instantiation parameter vector length and orientation, respectively.

\item For the brain graph capsule representation updating, we propose to involve the residual learning into the model, which can make the model much more robust in the learning process. The brain graph residual terms will be integrated into the class-wise digit capsule representation effectively.

\item We learn the {\our} model with the error back propagation algorithm. In addition to minimizing the model classification loss, we also include the raw brain graph reconstruction loss as a regularization task, whose introduced loss term will also involved for model learning.

\item We test the effectiveness of {\our} on several brain graph benchmark datasets and compare {\our} against both the classic and state-of-the-art brain graph representation learning models.

\end{itemize}

The remaining parts of this paper will be organized as follows. In Section~\ref{sec:related_work}, we will talk about the related research work. The terminology definitions and the problem formulation will be provided in Section~\ref{sec:formulate}. Detailed information about the proposed method will be introduced in Section~\ref{sec:method}, whose effectiveness will be tested in Section~\ref{sec:experiment}. Finally, we will conclude this paper in Section~\ref{sec:conclusion}.

\section{Related Work}\label{sec:related_work}

In this section, besides the brain graph learning background introduced in Section~\ref{sec:introduction}, we will talk about several important research topics related to this paper, which include \textit{graph neural network}, \textit{graph instance representation learning} and \textit{capsule network}.

\noindent \textbf{Graph Neural Network}: Traditional deep models (assuming the training instances are i.i.d.) cannot be directly applied to the graph data due to the extensively connected structure of graph data. In recent years, graph neural networks have become a very popular research topic for effective graph representation learning \cite{MBMRSB16,AT16,MBBV15,Kipf_Semi_CORR_16,SGTHM09,ZCZYLS18,NAK16}. GCN proposed in \cite{Kipf_Semi_CORR_16} feeds the generalized spectral features into the convolutional layer for representation learning. Similar to GCN, deep loopy graph neural network \cite{loopynet} proposes to update the node states in a synchronous manner, and it introduces a spanning tree based learning algorithm for training the model, which will not suffer from the suspended animation problem according to \cite{Zhang_GResNet_19}. GAT \cite{Velickovic_Graph_ICLR_18} leverages masked self-attentional layers to address the shortcomings of GCN. In recent years, we have also observed several derived variants of GCN and GAT models, e.g., \cite{Klicpera_Personalized_18,Li_Combinatorial_18,Gao_GraphNAS_19}. These new models further improves GCN and GAT in different perspectives.

\noindent \textbf{Graph Instance Representation Learning}: When it comes to the graph instance representation learning, some papers \cite{sdbn, Zhang2018AnED} propose to adopt the classic deep learning model, like CNN, as the encoder for embedding graph instances. To eliminate the order effects on the learned representations, SDBN \cite{sdbn} introduces a method to re-order the nodes in the graph instances. The DGCNN model \cite{Zhang2018AnED} accepts graphs of arbitrary structures, which can sequentially read a graph in a meaningful and consistent order. Some graph neural networks have also been proposed in recent years for learning graph instance representations \cite{gin, Ying2018HierarchicalGR, Meng_Isomorphic_NIPS_19}. The GIN model proposed in \cite{gin} can classify graph instances with node features. Meanwhile, the DIFFPOOL model \cite{Ying2018HierarchicalGR} has a differentiable graph pooling module that can generate hierarchical representations of graphs. The recent {\isonn} model \cite{Meng_Isomorphic_NIPS_19} learns a set of sub-graph templates for extracting the representations of input graph instances.

\noindent \textbf{Capsule Networks}: The capsule network initially proposed in \cite{NIPS2017_6975} has also attracted lots of attention from both researchers and practitioners in recent years. Capsule network has been demonstrated to be effective in various computer vision \cite{lalonde2018capsules} and natural language processing \cite{Towards_Capsule,zero_shot,multi_label_capsule} tasks. In computer vision, \cite{lalonde2018capsules} proposes to apply capsule network for object segmentation; based on brain raw imaging data, \cite{Capsule_Tumor_Classification} applies capsule network for brain tumor classification; and \cite{singh2019dual} studies the low-resolution image recognition with a dual directed capsule network. In NLP, \cite{Towards_Capsule} studies the application of capsule network in various natural language processing tasks, e.g., text classification and Q\&A; \cite{multi_label_capsule} studies the relation extraction with capsule network from textual data; and \cite{zero_shot} studies the zero-shot user intent detection task based on capsule network. Besides these application works, many research efforts have also been observed to improve capsule network from several different perspectives. \cite{NIPS2019_8982} proposes a self-routing strategy to address the high time cost and strong cluster assumptions of the input data; \cite{Deep_Capsule_Net} investigates to learning deep capsule network based on a novel 3D convolution based dynamic routing algorithm; whereas \cite{e2018matrix} interprets the dynamic routing process in capsule network from the EM perspective instead.


\section{Problem Formulation}\label{sec:formulate}

In this part, we will first introduce the notations to be used in this paper. After that, we will provide the definitions of several important terminologies and the formulation of the studied problem.

\subsection{General Notations}

In the sequel of this paper, we will use the lower case letters (e.g., $x$) to represent scalars, lower case bold letters (e.g., $\mb{x}$) to denote column vectors, bold-face upper case letters (e.g., $\mb{X}$) to denote matrices, and upper case calligraphic letters (e.g., $\mathcal{X}$) to denote sets or high-order tensors. Given a matrix $\mb{X}$, we denote $\mb{X}(i,:)$ and $\mb{X}(:,j)$ as its $i_{th}$ row and $j_{th}$ column, respectively. The ($i_{th}$, $j_{th}$) entry of matrix $\mb{X}$ can be denoted as $\mb{X}(i,j)$. We use $\mb{X}^\top$ and $\mb{x}^\top$ to represent the transpose of matrix $\mb{X}$ and vector $\mb{x}$. For vector $\mb{x}$, we represent its $L_p$-norm as $\left\| \mb{x} \right\|_p = (\sum_i |\mb{x}(i)|^p)^{\frac{1}{p}}$. The Frobenius-norm of matrix $\mb{X}$ is represented as $\left\| \mb{X} \right\|_F = (\sum_{i,j} |\mb{X}(i,j)|^2)^{\frac{1}{2}}$. The element-wise product of vectors $\mb{x}$ and $\mb{y}$ of the same dimension is represented as $\mb{x} \otimes \mb{y}$, whose concatenation is represented as $\mb{x} \sqcup \mb{y}$.

\begin{figure*}[t]
    \begin{minipage}{\textwidth}
    \centering

    	\includegraphics[width=0.8\linewidth]{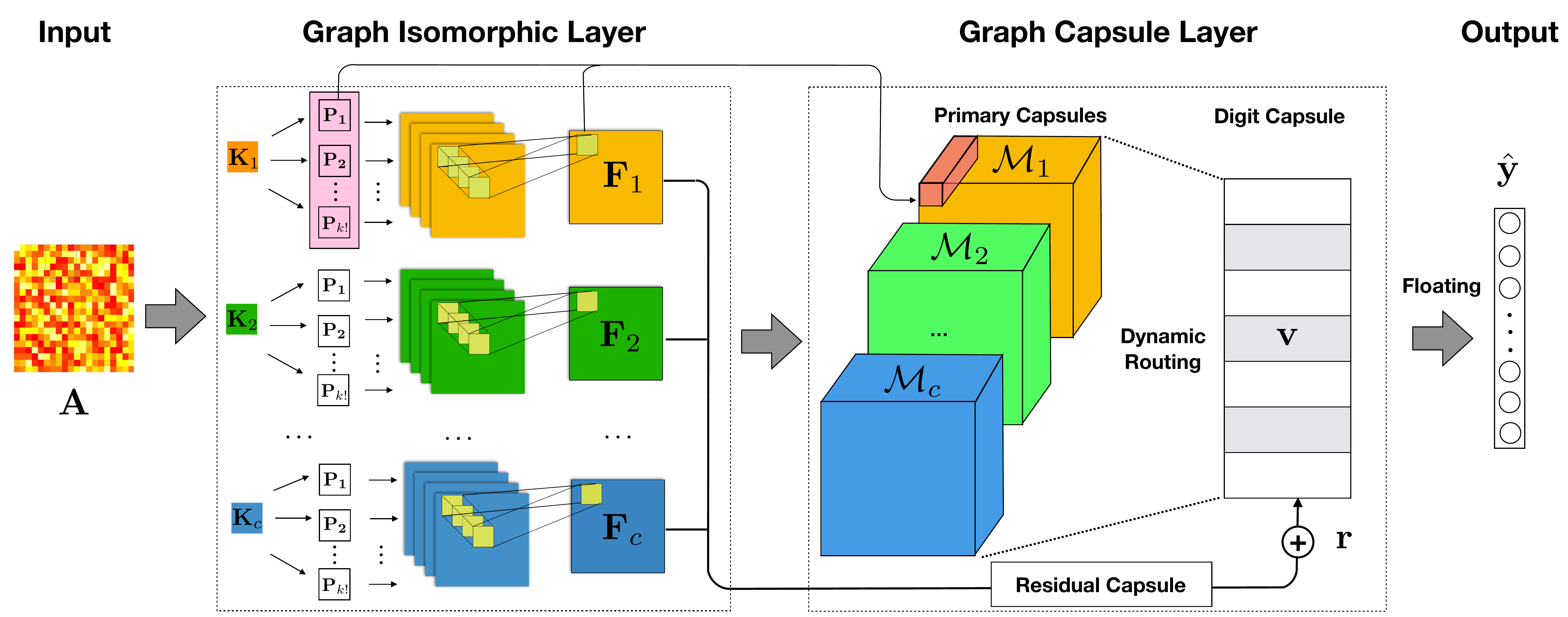}

    	\caption{An Illustration of the {\our} Framework for Brain Graph Representation Learning. The model has three main layers: (a) \textit{graph isomorphic layer}, (b) \textit{primary capsule layer}, and (c) \textit{classwise digit capsule layer}. With both the learned graph isomorphic features and the sub-graph templates, we will construct the primary capsules. The classwise digit capsule can be effectively learned via the dynamic routing process together with the residual capsules term. Via the label floating process, we will project the classwise digit capsule to the corresponding label vector of the input brain graph instance.}
    	\label{fig:architecture}
    \end{minipage}%

\end{figure*}

\subsection{Terminology Definitions}

Here, we will provide the definitions of several important terminologies used in this paper, which include \textit{brain graph instance}, \textit{adjacency matrix} and \textit{brain graph set}.

\begin{definition}
(Brain Graph Instance): Formally, a brain graph instance can be denoted as $G=(\mc{V}, \mc{E}, w)$, where $\mc{V}$ and $\mc{E}$ denote the sets of nodes and links in the graph, respectively. Mapping $w: \mc{E} \to \mathbbm{R}$ indicates the weights of links in the brain graph instance. For unweighted brain graphs, we will have $w(e_{i,j}) = 1, \forall e_{i,j} \in \mc{E}$ and $w(e_{i,j}) = 0, \forall e_{i,j} \in \mc{V} \times \mc{V} \setminus \mc{E}$. Here, $e_{i,j} = (v_i, v_j)$ denotes the link between node pair $v_i$ and $v_j$ in the brain graph instance $G$.
\end{definition}

The brain cerebral cortex can be partition into different areas according to their functions, locations and cortex layers. There exist different brain area partitions, such as the \textit{Brodmann areas} and the \textit{automated anatomical labeling} (AAL). Each node in the above brain graph definition corresponds to a specific area of human brain, while the edges in the brain graph indicate their activity correlations in the scanned brain image.

\begin{definition}
(Brain Graph Matrix): Given an input brain graph instance $G$, we can denote its structure as an adjacency matrix $\mb{A} \in \mathbbm{R}^{|\mc{V}| \times |\mc{V}|}$, where entry $\mb{A}(i,j) = w(e_{i,j})$ denotes the weight of the link $e_{i,j} = (v_i, v_j)$ in the input brain graph.
\end{definition}

Different from the graph data studied in other areas, since the nodes in the graph correspond to specific cerebral cortex area, there exists an inherent order among the nodes in the graph. By default, all the brain graphs will have the identical node orders when defining their corresponding adjacency matrices according to the brain area partition algorithm.

\begin{definition}
(Brain Graph Set): For each brain graph instance studied in this paper, it can be attached with some pre-defined class labels. Formally, we can represent $n$ labeled graph instances studied in this paper as a set $\mc{G} = \{(G_i, {y}_i)\}_{i = 1}^n$, where ${y}_i \in \mc{Y}$ denotes the class label of brain graph $G_i$ and $\mc{Y}$ is the label space. For presentation simplicity, in reference to the studied graph set (without labels), we can also denote them as set notation $\mc{G}$ as well, which will be used in the following problem formulation.
\end{definition}

\subsection{Problem Formulation}

Based on the above notations and terminologies, we can provide the formal definition of the problem studied in this paper as follows.

\noindent \textbf{Problem Definition}: Formally, given the partially labeled brain graph instances in set $\mc{G}$, we aim at learning a mapping $f: \mc{G} \to \mathbbm{R}^{d}$ to project the brain graph instances to their corresponding feature representations with dimension $d$. Here, we also add extra requirements on mapping $f$, which should be able to learn node-order and view-point invariant representations of the brain graph instances. In addition, in the learned representations, it should capture not only the existence information but also the orientation properties of the sub-graph patterns in the brain graph instances, which will be fed to the downstream functional component to further infer the class labels for brain graph classification.


\section{Proposed Method}\label{sec:method}

In this section, we will introduce the {\our} model for brain graph representation learning and classification. As illustrated in Figure~\ref{fig:architecture}, given the input brain graph instance $G \in \mc{G}$, we can represent its structure as its adjacency matrix $\mb{A}$. The {\our} model can effectively learn the representation for the brain graph instance based on $\mb{A}$ via several layers: (a) \textit{graph isomorphic layer}, (b) \textit{primary capsule layer}, and (c) \textit{class-wise digit capsule layer}, respectively. The detailed functionality and learning mechanism of these different layers will be illustrated in the following parts of this section.


\subsection{Graph Isomorphic Layer}\label{subsec:isomorphic_layer}

The graph isomorphic layer involved in {\our} aims to learn the node-order invariant isomorphic representations for the input brain graph instance. By learning a set of sub-graph templates, {\our} will match the input brain graph instances with the templates for learning their representations. Such learned representations can be used to initialize the primary capsules to be introduced in the following subsection.

Formally, given an input brain graph $G$, {\our} introduces a set of sub-graph template variables, i.e., $\{\mb{K}_1, \mb{K}_2, \cdots, \mb{K}_c\}$ and $\mb{K}_i \in \mathbbm{R}^{k \times k}$ ($c$ and $k$ denotes the channel number and sub-graph template size, respectively), to extract the sub-graph patters from the input brain graph instance. Instead of enumerating all the potential combinations of $k$ different nodes from the input brain graphs, by shifting the template $\mb{K}_i$ along the dimensions of the input brain graph denoted by its adjacency matrix $\mb{A}$, {\our} can compute the isomorphic feature for the regional sub-matrix $\mb{A}_{s,t}$ (i.e., $\mb{A}(s\text{:}s+k-1, t\text{:}t+k-1)$ and $s,t$ denote the row/column index pair) by matching $\mb{K}_i$ with $\mb{A}_{s,t}$. The introduced matching scores will be defined as the isomorphic feature for regional sub-matrix $\mb{A}_{s,t}$ based on template $\mb{K}_i$ shown as follows:
\begin{equation}\label{equ:isomorphic_feature}
\mb{F}_i(s,t) =1.0 - \min_{\mb{P} \in \mc{S}} \left\| \mb{P} \mb{K}_i \mb{P}^\top - \mb{A}_{s,t} \right\|_F,
\end{equation}
where $\mc{S}$ denotes the space of $k!$ potential permutation matrices. Each permutation matrix $\mb{P} \in \mc{S}$ will permutate the node orders to find the optimal matching between the regional sub-matrix and the template matrix. A perfect matching between the permutation matrix and regional sub-matrix will lead to $0$ matching loss, i.e., the learned feature will be $1.0$ according to the above equation. All such extracted features by template $\mb{K}_i$ on different regional sub-matrices of $\mb{A}$ will be organized as matrix $\mb{F}_i \in \mathbbm{R}^{(|\mc{V}|-k+1) \times (|\mc{V}|-k+1)}$.

The original isomorphic feature extraction approach introduced in {\isonn} \cite{Meng_Isomorphic_NIPS_19} can be very costly, since the size of potential permutation matrix set, i.e., $\mc{P}$, is $k!$, which will increase exponentially as $k$ goes up. To lower down the time cost, the optimal permutation matrix $\mb{P}$ can also be calculated directly for a fast brain graph isomorphic feature extraction. Formally, let the kernel variable $\mb{K}_i$ and the input regional sub-matrix $\mb{A}_{s,t}$ be $k\times k$ real symmetric matrices with $k$ distinct eigenvalues $\alpha_1 > \alpha_2 > \cdots > \alpha_k$ and $\beta_1 > \beta_2 > \cdots > \beta_k$, respectively, and their eigendecomposition be represented by 
\begin{equation}\label{eq:eigen}
\mb{K}_i = \mb{U}_{1}\mb{\Lambda}_{1}\mb{U}_{1}^{\top}, \mbox{ and } \mb{A}_{s,t} =\mb{U}_{2}\mb{\Lambda}_{2}\mb{U}_{2}^{\top},
\end{equation} 
where $\mb{U}_{1} $ and $\mb{U}_{2}$ are orthogonal matrices of eigenvectors and $\mb{\Lambda}_{1}= diag(\alpha_1, \cdots, \alpha_k), \mb{\Lambda}_{2}=diag(\beta_1, \cdots, \beta_k)$. The minimum of $||\mb{P}\mb{K}_i\mb{P}^{\top}-\mb{A}_{s,t}||^2$ can actually be attained with the following $\mb{P}^*$:
\begin{equation}
\mb{P}^* = \mb{U}_{2}\mb{S}\mb{U}_{1}^{\top},
\end{equation} 
where $\mb{S} \in \mathcal{S} = \{diag(s_1,s_2,\cdots, s_k)|s_i=1 \mbox{  or }  -1 \}$.

The above approach allows a fast isomorphic feature extraction of brain graph instances. Meanwhile, different from {\isonn} \cite{Meng_Isomorphic_NIPS_19}, which only uses $\mb{F}_i$ as the output of the isomorphic layer, to clearly indicate the sub-graph pose and orientation properties, we will also extract the learned optimal permutation matrix $\mb{P}$ as one part of the learned representations. Formally, the corresponding matrix $\mb{P} \in \mc{S}$ that achieve $\mb{F}_i(s,t)$ in Equation~(\ref{equ:isomorphic_feature}) based on the sub-graph template $\mb{K}_i$ can be denoted as $\mb{P}_{i,s,t}$ (or its reshaped vector $\mb{p}_{i,s,t} \in \mathbbm{R}^{k^2}$ for representation simplicity). For all the regional sub-matrices in $\mb{A}$, its corresponding optimal permutation matrices can be organized into a high-way tensor $\mc{P}_i \in \mathbbm{R}^{(|\mc{V}|-k+1) \times (|\mc{V}|-k+1) \times k^2}$, where $\mc{P}_i(s,t,:) = \mb{p}_{i,s,t}$. The above operations can be applied to all the sub-graph templates $\mb{K}_1$, $\mb{K}_2$, $\cdots$, $\mb{K}_c$. For the other sub-graph templates, we can denote their extracted brain graph isomorphic feature matrices and optimal permutation tensors as $\left\{\mb{F}_1, \mb{F}_2, \cdots, \mb{F}_c \right\}$ and $\left\{\mc{P}_1, \mc{P}_2, \cdots, \mc{P}_c \right\}$, respectively, which will be fed to initialize the brain graph primary capsules in the following subsection.


\subsection{Brain Graph Primary Capsule Layer}

According to the previous description, the learned feature matrix $\mb{F}_i$ delivers the existence confidence scores of sub-graphs in the input brain graph instance. Meanwhile, the learned high-way tensor $\mc{P}_i$ constructed by the permutation matrices (or reshaped vectors) can effectively illustrate the pose, size and orientation properties of such regional sub-graphs. Existing graph isomorphic representation learning methods, like {\isonn} \cite{Meng_Isomorphic_NIPS_19}, merely use the learned feature representation $\mb{F}_i$ to classify the graph instances, where the sub-graph pose information, i.e., tensor $\mc{P}_i$, is discarded. Here, we propose to construct the primary capsule based on both $\mb{F}_i$ and $\mc{P}_i$ to involve both the sub-graph existence confidence scores and the sub-graph pose property information in the learned representations.

Formally, we can denote the primary capsule component constructed based on $\mb{F}_i$ and $\mc{P}_i$ as $\mc{M}_i \in \mathbbm{R}^{(|\mc{V}|-k+1) \times (|\mc{V}|-k+1) \times d_m}$, where $d_m$ denotes the primary capsule instantiated parameter vector length. The entries, e.g., $\mc{M}_i(s,t, :)$, are defined based on both $\mb{F}_i$ and $\mc{P}_i$, which can be formally represented as follows:
\begin{equation}\label{equ:primary_capsule}
\mc{M}_i(s,t, :) = \frac{g\left( \mc{P}_i(s,t,:) \right)}{\left\| g\left( \mc{P}_i(s,t,:) \right) \right\|} \cdot \mb{F}_i(s,t).
\end{equation}
In the above equation, $\left\| \cdot \right\|$ denotes the input vector length for normalization. Function $g(\cdot)$ is a dimension adjustment mapping, which can help shrink the capsule instantiation vector size from $k^2$ to $d_m$ (in the case if $k^2$ is large). In this paper, we  will just have $g(\mb{x}) = \mb{x}$ for simplicity, as we tend to use small-sized sub-graph templates, so the $k^2$ value will still be acceptable.

Based on the above representations, it is easy to observe that the primary capsule $\mc{M}_i$ illustrates both the existence scores and basic properties of sub-graph patterns in the input brain graph:
\begin{itemize}

\item The length of $\mc{M}_i(s,t, :)$ denotes the existence confidence score of sub-graph template $\mb{K}_i$ in the regional input brain graph, whose value equals to $\mb{F}_i(s,t)$.

\item The orientation and dimensionality of $\mc{M}_i(s,t, :)$ can effective indicate the pose, orientation and size of the sub-graph in the regional input brain graph, where the orientation can be denoted by the normalized unit vector $\frac{g\left( \mc{P}_i(s,t,:) \right)}{\left\| g\left( \mc{P}_i(s,t,:) \right) \right\|}$. 

\end{itemize}

For the other sub-graph templates in $\{\mb{K}_1, \mb{K}_2, \cdots, \mb{K}_c\}$, we can also construct the corresponding tensors in a similar way, which together can be denoted as a four-way tensor 
\begin{equation}
\mc{M} = [\mc{M}_1, \mc{M}_2, \cdots, \mc{M}_c] \in \mathbbm{R}^{c \times (|\mc{V}|-k+1) \times (|\mc{V}|-k+1) \times d_m}.
\end{equation}
We have $d_m = k^2$ in this paper for simplcity. Tensor $\mc{M}$ will serve as the primary capsule for representation updating with the dynamic routing algorithm to learn the class-wise digit capsule to be introduced in the following subsections.


\subsection{Class-wise Digit Capsule Layer}

Based on the learned sub-graph property primary capsule, we can further compute the class-wise digit capsule to represent the brain graph class label inference results. Formally, we can denote such digit capsule as $\mb{C} \in \mathbbm{R}^{|\mc{Y}| \times d_c}$, where $d_c$ denotes the size of the class-wise instantiated parameter vector. To be more specific, for any class $y_j \in \mc{Y}$, we can denote the corresponding instantiated vector learned for the input brain graph as $\mb{C}(j,:)$ (or $\mb{c}_j$ for simplicity). 

The class-wise digit capsule vector can be computed with the brain graph primary capsule defined above effectively. Formally, the primary capsule $\mc{M}$ can be treated as a group of $c \times (|\mc{V}|-k+1) \times (|\mc{V}|-k+1)$ instantiated vector of dimension $d_m$, which will compose the class-wise instantiated parameter vector $\mb{c}_j$ corresponding to class label $y_j \in \mc{Y}$ according to the following equation:
\begin{equation}\label{equ:digit_capsule}
\begin{aligned}
&\mb{c}_j = \mbox{squash}(\bar{\mb{c}}_j ) = \frac{\left\| \bar{\mb{c}}_j \right\|^2}{1+\left\| \bar{\mb{c}}_j \right\|^2} \frac{\bar{\mb{c}}_j}{\left\| \bar{\mb{c}}_j \right\|},\\
&\text{where } \bar{\mb{c}}_j = \sum_{i, s, t} \bs{\alpha}_j(i,s,t) \cdot  \text{MLP}(\mb{m}_{i,s,t}; \mb{W}).
\end{aligned}
\end{equation}
In the above equation, vector $\mb{m}_{i,s,t} = \text{Padding}(\mc{M}(i,s,t,:))$ and $\mc{M}(i,s,t,:)$ is exactly the same as the term we define in Equation~(\ref{equ:primary_capsule}). Meanwhile, considering that the permutation matrix we use for defining $\mc{M}(i,s,t,:)$ is extremely sparse (among the $k^2$ elements, $k^2-k$ of them are actually zeros), it will lead to defective capsule values after feeding it to the squash function. In this paper, we propose to padding the zero elements in $\mc{M}(i,s,t,:)$ with a default constant value $\gamma \in [-1, 1]$. Notation $\mbox{squash}(\cdot)$ denotes the squashing function, which also normalizes the input vectors to a length slightly below $1$. Term $\bs{\alpha}_j(i,s,t)$ is the weight between the primary capsule and the digit capsule, and weight matrix $\mb{W} \in \mathbbm{R}^{d_m \times d_c}$ is used for the vector dimension adjustment. The second formula used above actually aggregates the information from the primary capsule to reason the potential class-wise digit capsule.

Capsule network \cite{Sabour_Capsule_17} introduces a learning mechanism to update the connection weights $\bs{\alpha}_j$, which can be denoted as the following equations:
\begin{equation}
\begin{cases}
\mbox{Initialize: }\hspace{-10pt}& \bs{\beta}_j(i,s,t) = {0},\\
\mbox{Update: }\hspace{-10pt}& \bs{\beta}_j(i,s,t) = \bs{\beta}_j(i,s,t) + \mb{c}_j^\top \mb{m}_{i,s,t},\\
\mbox{Normalize: }\hspace{-10pt}& \bs{\alpha}_j = \mbox{leaky-softmax}(\bs{\beta}_j).
\end{cases}
\end{equation}
In the iterative updating process, vector $\mb{c}_j$ will be replaced with the latest class-wise instantiated parameter vector computed by Equation~(\ref{equ:digit_capsule}) defined above. Such a process is also named as dynamic routing in capsule network.


\subsection{Residual Capsule Layer}

In this paper, we also extend the capsule network model by introducing the residual capsule layer, which can also accept the brain graph intermediate representations in computing the model class-wise digit capsule introduced before. Formally, based on the learned brain graph isomorphic representation matrices $\mb{F}_1, \mb{F}_2, \cdots, \mb{F}_c$ with elements defined in Equation~(\ref{equ:isomorphic_feature}), we can also project such representations to a class digit capsule via several fully-connected layers:
\begin{equation}
\begin{aligned}
&\mb{F}_i = \mbox{softmax}\left( \mb{F}_i \right) \forall i \in \{1, 2, \cdots, c\},\\
&\mc{F} = [\mb{F}_1, \mb{F}_2, \cdots, \mb{F}_c],\\
&\mb{r}_j = \mbox{MLP}(\mbox{reshape}(\mc{F}); \mb{W}_{j}).
\end{aligned}
\end{equation}
In the above equations, term $\mb{r}_j$ denotes the capsule vector corresponding to the class label $y_j \in \mc{Y}$. Prior to feeding the learned representations to compute the residual capsules, we propose to normalize each learned representations with the softmax function on each $\mb{F}_i, \forall i \in \{1,2,\cdots,c\}$. Such normalized representations will be grouped into a tensor $\mc{F}$, which will be reshaped into different class capsule vectors for classification with several fully-connected layers. Here, vector $\mb{r}_j$ is of the same dimension as the class digit capsule $\mb{c}_j$ learned in the previous subsection, which together can define the final class representations:
\begin{equation}
\mb{v}_j = \mb{r}_j + \mb{c}_j, \forall y_j \in \mc{Y}.
\end{equation}
Compared with the vanilla capsule network, assisted with the residual capsule layer, the model's performance can be much more stable and robust in the dynamic routing process.

\subsection{Framework Learning}


In {\our}, we use the length of the instantiation vectors to denote the probability that a capsule's entity exists. Formally, given the input brain graph instance $G_g \in \mc{G}$, its learned class-wise dight capsule vectors can be denoted as $\{\mb{v}_0, \mb{v}_1, \cdots, \mb{v}_{|\mc{Y}|-1}\}$, respectively. For the binary classification task studied in this paper, we will have two digit capsule vectors $\mb{v}_{0}, \mb{v}_{1}$ for the negative and positive classes. For all these digit capsule, compared against its ground truth label, we can define the introduced loss term as follows:
\begin{equation}
\begin{aligned}
\ell(\mc{G}) = \sum_{G_g \in \mc{G}} \sum_{y_j \in \mc{Y}}& \Big( t_j \cdot \max(0, 0.9 - \left\| \mb{v}_j \right\|)^2 \\
&+ 0.5 \cdot (1- t_j) \cdot \max(0, \left\| \mb{v}_j  \right\| - 0.1)^2 \Big),
\end{aligned}
\end{equation}
where the temporary binary indicator term $t_j = 1$ denotes the true class label for the graph instance is $y_j \in \mc{Y}$.


In addition to the classification loss function, in this paper, we also introduce a reconstruction component to recover the brain graph adjacency matrix $\mb{A}_g$ for brain graph $G_g \in \mc{G}$ based on the learned class-wise dight capsule vectors. Such a component can maintain the quality of the learned brain graph capsule representations, and also avoid some trivial solutions. Formally, the brain graph reconstruction is achieved with several fully-connected layers, which can be denoted as follows:
\begin{equation}
\hat{\mb{A}}_g = \mbox{reshape} \left( \mbox{MLP} \left(\mb{v}_1 \sqcup \mb{v}_2 \sqcup \cdots \sqcup \mb{v}_{|\mc{Y}|}; \mb{W} \right) \right).
\end{equation}
The $\mbox{reshape}(\cdot)$ function used here will transform the output vector into a matrix shape and $\hat{\mb{A}}_g \in \mathbbm{R}^{|\mc{V}| \times |\mc{V}|}$ and the operator $\mb{v}_1 \sqcup \mb{v}_2$ will concatenate the vectors. Formally the regularization loss introduced for the brain graph reconstruction can be denoted as
\begin{equation}
\ell_{reg}(\mc{G}) =  \sum_{G_g \in \mc{G}} \left\| \mb{A}_g - \hat{\mb{A}}_g \right\|_F.
\end{equation}

Since the graph reconstruction loss term value scale is much larger than the label inference loss. In this paper, we propose to balance between these two loss term and define the final learning objective function as follows
\begin{equation}
\ell (\mc{G}) = \frac{1}{|\mc{G}|} \left( \ell(\mc{G}) + \delta \cdot \ell_{reg}(\mc{G}) \right).
\end{equation}
According to our experimental tests on the dataset, the scale weight term $\delta = 0.0005$ works the best for the the {\our} in terms of the label inference results. By simultaneously minimizing both the loss function and the regularization term, we will be able to learn both the model variables and the brain graph representations.

\section{Experiments}\label{sec:experiment}

To test the effectiveness of the proposed model, we have conducted extensive experiments on several real-world brain graph benchmark datasets. In this section, we will first briefly describe the datasets used in this paper. After that, we will introduce the experimental settings and provide the experimental results with detailed analyses.

\subsection{Dataset Description}

\noindent \textbf{Datasets}: The datasets used in the experiments include several brain graph benchmark datasets, HIV-fMRI, HIV-DTI, BP-fMRI and ADHD-fMRI. These datasets are extensively used in the existing brain graph analysis and classification research papers \cite{Meng_Isomorphic_NIPS_19, Lee2017IdentifyingDC, doi:10.1137/1.9781611974973.21, identifying_brain}. All these graph instances in the datasets have been labeled as either positive or negative instances by domain experts, which denote the sick patients and the seronegative control subjects. Via the {\our} model proposed in this paper and other comparison methods, we aim to learn the representations of these graph instances, which will be further used to infer their classification labels. Along with the brain graph classification task, {\our} can also reconstruct the input brain graphs as well, whose performance will also be investigated in the experiments as well.

\noindent \textbf{Source Code for Reproducibility}: Both the datasets and source code used in this paper are publicized for experimental results reproducibility, which can be accessed via link\footnote{https://github.com/jwzhanggy/IsoCapsNet} at the footnote.

\subsection{Experimental Settings}

\subsubsection{Experimental Setups}

For the {\our} method proposed in this paper, based on the input graph isomorphic features and permutation matrices learned by the isomorphic layer, we will construct the primary capsule. Furthermore, via the dynamic routing, we will also calculate the class-wise digit capsule for both the classification task and input graph reconstruction. For fair comparison, similar to the existing papers \cite{Meng_Isomorphic_NIPS_19, identifying_brain}, we partition the graph datasets into training and testing sets via 3-fold cross-validation, with 2-fold as the training set and 1-fold as the testing set. We will report the average performance scores as the final results. To train the proposed model, in the experiments, we use Adam with learning rate $0.01$ and decay weight $5e-4$ as the default optimizer. The default maximum training epoch is set to be $100$ for all methods and an early stop or extra learning epochs will be used when necessary. In each training epoch, we will iterate through all the data instances with a mini-batch size $64$. For the {\our} method, we will use the default kernel size parameter $k=4$. Analyses of the hyper-parameter $k$ with different values will also be provided in the following part of the experiments. Necessary parameter tuning for different comparison methods will be performed when needed. All these methods are compared on the Dell PowerEdge T630 Server with a 20-core Intel CPUs, 256GB memory and 128 GB SSD swap.

\begin{table*}[t]
\vspace{-10pt}
\caption{Graph Classification Results on Benchmark Datasets Evaluated by Averaged Accuracy and F1 Scores. (For the method(s) with the best scores, we highlight their results with bolded fonts. Also the top-3 methods rankings are indicated by the ``{\footnotesize \color{blue}(1)} {\footnotesize \color{blue}(2)} {\footnotesize \color{blue}(3)}'' numbers within the parentheses attached next to the scores.)}
\label{tab:main_result}
\vspace{-10pt}
\centering
{\setlength{\tabcolsep}{2.0pt}
\begin{tabular}{cc | cc | cccc | ccc | cc | c | c}
\toprule
\multirow{3}{*}{\textbf{metric}}&\multirow{3}{*}{\textbf{dataset}}&\multicolumn{9}{|c|}{\textbf{Other Baseline Methods}}&\multicolumn{3}{|c|}{\textbf{Ablation Methods}} & \textbf{Our Method}\\
\cmidrule{3-15}
&&{\wl} &{\freq} &{\cnn} &{\sdbn} &{\dgcnn} &{\autoencoder} &{\gcn} &{\gin} &{\diffpool} &{\fastisonn} &{\isonn} &{\capsule}	&{\our}\\
\midrule
\multirow{5}{*}{\rotatebox{90}{Accuracy}}
&{HIV-fMRI} &0.530	&0.543	&0.593	&0.665	&0.707	&0.469	&0.583	&0.643	&0.679	&0.739	&0.765 {\footnotesize \color{blue}(2)}	&0.746 {\footnotesize \color{blue}(3)}	&\textbf{0.805} {\footnotesize \color{blue}(1)}\\
\cmidrule{2-15}
&{HIV-DTI} &0.427	&0.646	&0.543	&0.659	&0.487	&0.624	&0.624	&0.626	&0.650	&0.601	&0.675 {\footnotesize \color{blue}(3)}	&0.731 {\footnotesize \color{blue}(2)}	&\textbf{0.750} {\footnotesize \color{blue}(1)}\\
\cmidrule{2-15}
&{BP-fMRI} &0.545	&0.568	&0.546	&0.648 {\footnotesize \color{blue}(3)}	&0.639	&0.536	&0.609	&0.577	&0.619	&0.623	&0.649 {\footnotesize \color{blue}(2)}	&0.576	&\textbf{0.669} {\footnotesize \color{blue}(1)}\\
\cmidrule{2-15}
&{ADHD-fMRI} &0.502	&0.551	&0.569	&0.503	&0.614	&0.562	&0.581	&0.530	&0.607	&0.624 {\footnotesize \color{blue}(2)}	&\textbf{0.641} {\footnotesize \color{blue}(1)}	&0.563	&0.621 {\footnotesize \color{blue}(3)}\\
\midrule

\midrule
\multirow{5}{*}{\rotatebox{90}{F1}}
&{HIV-fMRI} &0.451	&0.593	&0.663	&0.667	&0.688	&0.355	&0.625	&0.638	&0.684	&0.707 {\footnotesize \color{blue}(3)}	&0.759 {\footnotesize \color{blue}(2)}	&0.700	&\textbf{0.801} {\footnotesize \color{blue}(1)}\\
\cmidrule{2-15}
&{HIV-DTI} &0.407	&0.639 {\footnotesize \color{blue}(2)}	&0.557	&0.656	&0.442	&0.589 &0.634	&0.627	&0.616	&0.619	&\textbf{0.723} {\footnotesize \color{blue}(1)}	&0.617	&0.672 {\footnotesize \color{blue}(3)}\\
\cmidrule{2-15}
&{BP-fMRI} &0.598	&0.576	&0.528	&0.637	&0.671	&0.695 {\footnotesize \color{blue}(3)}	&0.643	&0.603	&0.654	&0.632	&0.697 {\footnotesize \color{blue}(2)}	&0.547	&\textbf{0.703} {\footnotesize \color{blue}(1)}\\
\cmidrule{2-15}
&{ADHD-fMRI} &0.475	&0.554	&0.572	&0.668 {\footnotesize \color{blue}(2)}	&0.616	&0.667 {\footnotesize \color{blue}(3)}	&0.593	&0.595	&0.619	&0.623	&{0.656}	&0.594	&\textbf{0.671} {\footnotesize \color{blue}(1)}\\

\bottomrule
\end{tabular}
}
\vspace{-10pt}
\end{table*}

\subsubsection{Comparison Methods}

The graph classification methods compared in the experiments can be divided into five categories, which include both the \textit{classic graph learning methods}, \textit{conventional representation learning methods}, \textit{graph neural network methods}, \textit{latest deep models} related to this paper and our proposed method.

\noindent \textit{Conventional Graph Learning Methods}:
\begin{itemize}
\item \textbf{Freq}: The {\freq} method \cite{1184038} is a classic graph classification algorithm, which extracts the top-k frequent sub-graphs as the features. The feature extraction process is purely based on the frequency and works in an unsupervised manner.

\item \textbf{WL}: The {\wl} method \cite{NIPS2009_0a49e3c3} classifies graph instances in the testing set into different classes based on the weighted summation of labels of graph instances in training set subject to the graph isomorphism tests.
\end{itemize}

\noindent \textit{Conventional Representation Learning Methods}:
\begin{itemize}
\item \textbf{CNN}: Given the graph input, we can represent the graph structure as its adjacency matrix. In the experiments, we also apply the {\cnn} model \cite{10.5555/646469.691875} with the LeNet architecture for learning the graph representations.
\item \textbf{SDBN}: To eliminate the order effects on the learned representations, {\sdbn} \cite{sdbn} introduces a method to re-ordered the nodes in the graph prior to feeding them to {\cnn} for learning their representations and classification labels.
\item \textbf{DGCNN}: The {\dgcnn} model \cite{Zhang2018AnED} accepts graphs of arbitrary structures. Besides extracting useful features characterizing the rich information encoded in a graph, {\dgcnn} can also sequentially read a graph in a consistent order.
\item \textbf{Autoencoder}: Via the {\autoencoder} model \cite{10.5555/1756006.1953039}, graphs can be compressed into low-dimensional embedding vectors, which will be used to train a MLP for classification purposes.
\end{itemize}

\noindent \textit{Latest Graph Representation Learning Methods}:
\begin{itemize}
\item \textbf{GCN}: Besides the {\cnn} model, we also include {\gcn} \cite{Kipf_Semi_CORR_16} as a comparison methods in the experiments. With the graph convolutional operator, we aggregate node representations as the embedding of the whole graph instance.
\item \textbf{GIN}: The {\gin} model proposed in \cite{gin} can classify graph instances with node features. In this experiment, we also include {\gin} as a comparison baseline method.
\item \textbf{DIFFPOOL}: The {\diffpool} model \cite{Ying2018HierarchicalGR} has a differentiable graph pooling module that can generate hierarchical representations of graphs and can be combined with various graph neural network architectures in an end-to-end fashion. 
\end{itemize}

\noindent \textit{Deep Learning Methods Related to This Paper}:
\begin{itemize}
\item \textbf{IsoNN}: The {\isonn} model \cite{Meng_Isomorphic_NIPS_19} learns a set of sub-graph templates for extract the representations of input graph instances. The {\isonn} model cannot consider the orientation of the sub-graph templates in the learned representations.
\item \textbf{FastIsoNN}: Instead of enumerating the permutation matrices, as introduced in Section~\ref{subsec:isomorphic_layer} of this paper, the {\fastisonn} model \cite{Meng_Isomorphic_NIPS_19} is a fast version of {\isonn} and calculates the optimal permutation matrix instead.
\item \textbf{CapsNet}: Via the dynamic routing function, the {\capsule} model \cite{Sabour_Capsule_17} can also learn the graph instance representations as capsules. In the experiment, {\capsule} uses {\cnn} for constructing the primary capsules.
\end{itemize}

\noindent \textit{Methods Proposed in This Paper}:
\begin{itemize}
\item \textbf{\our}: The {\our} model proposed in this paper proposes to construct the primary capsule with both the learned sub-graph templates and the sub-graph existence confidence scores. The graph instances will be further compressed into embedding representations via the dynamic routing function for classification.
\end{itemize}

\subsubsection{Evaluation Metrics}

The classification results obtained by all the comparison methods will be evaluated by both Accuracy and F1 as the evaluation metrics. To be consistent with the existing papers, the average scores obtained in all the cross-validation folds will be reported. For some performance analysis of the methods, we will also use the learning time costs as an extra evaluation metric.

\begin{table*}[t]
\vspace{-10pt}
\caption{Analysis of the impacts of sub-graph size on {\our} model learning performance. In the table, $k$ denotes the sub-graph size, the results are evaluated by both Accuracy and F1. The time (in seconds) denotes time costs for model training. The last column denotes the weighted sub-graph kernel learned by the model.}
\vspace{-5pt}
\label{tab:parameter_analysis}
\centering
{\setlength{\tabcolsep}{8pt}
\begin{tabular}{cccccc}
\toprule
k & Acccuracy & F1 & Time(s) & Learned Template Matrix & Learned Sub-graph Structure\\
\midrule
1 & 0.261 & 0.108 & 3.579 & 
$\mb{K} = \begin{bmatrix}
0.2225\\
\end{bmatrix}$
&\begin{minipage}{.05\textwidth}
      \includegraphics[width=\linewidth]{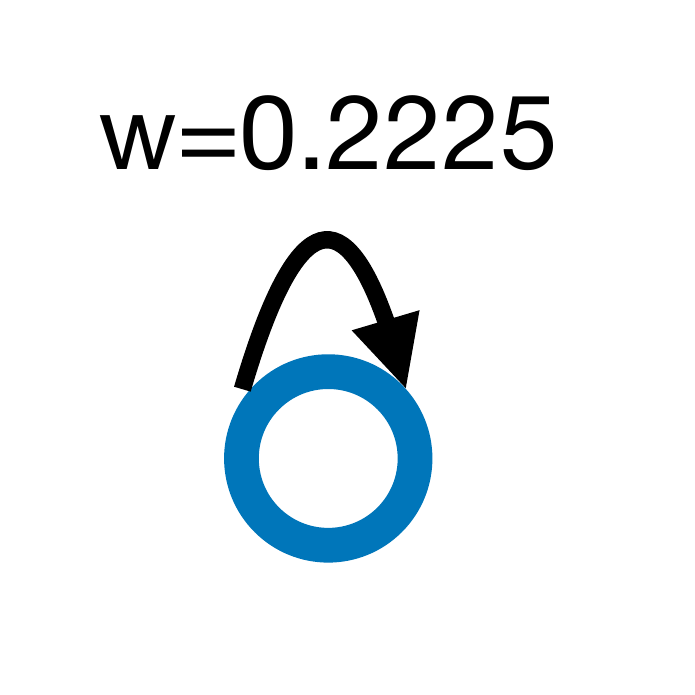}
    \end{minipage} \\
\midrule
2 & 0.435 & 0.448 & 4.862 &
$\mb{K} = \begin{bmatrix}
0.2181 & -0.3791\\
-0.4737 &-1.3556\\
\end{bmatrix}$ 
&\begin{minipage}{.07\textwidth}
      \includegraphics[width=\linewidth]{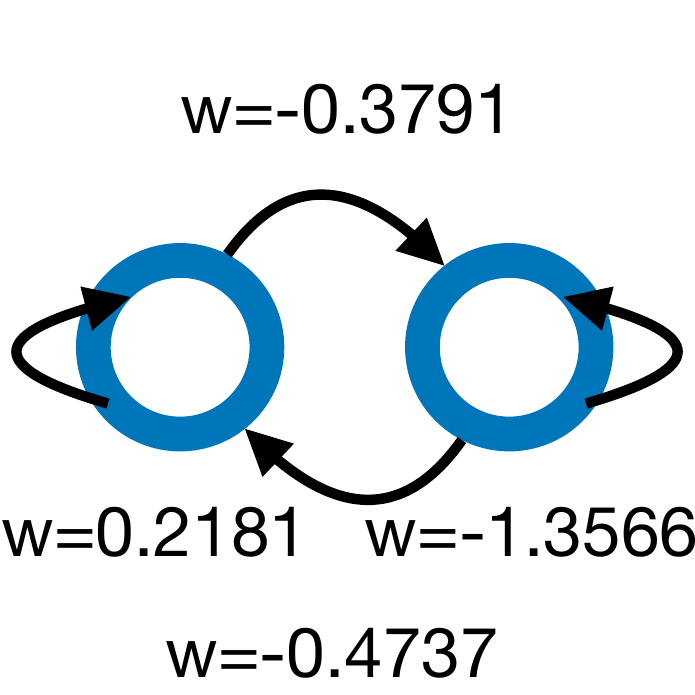}
    \end{minipage} \\
\midrule
3 & 0.696 & 0.606 & 6.186 &
$\mb{K} = \begin{bmatrix}
0.4356 & -0.3431 & -0.4322\\
-1.3110 & 1.1301 & -0.7403\\
0.6213 & 0.4688 & -0.5037\\
\end{bmatrix}$ 
&\begin{minipage}{.09\textwidth}
      \includegraphics[width=\linewidth]{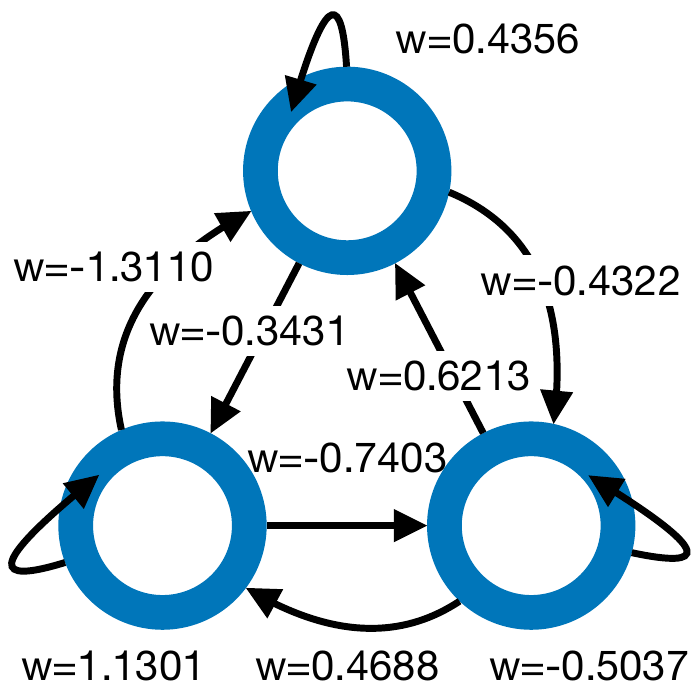}
    \end{minipage} \\
\midrule
4 & \textbf{0.805} & \textbf{0.801} & 10.995 &
$\mb{K} = \begin{bmatrix}
-0.8729 & 0.7488 & -1.1359 & -0.9417\\
-1.0507 & 0.9993 & -0.8895 & 0.0180\\
-0.7403 & 0.1812 & -0.4052 & -0.0766\\
0.8396 & -0.5859 & -0.0670 & -0.2680\\
\end{bmatrix}$ 
&\begin{minipage}{.13\textwidth}
      \includegraphics[width=\linewidth]{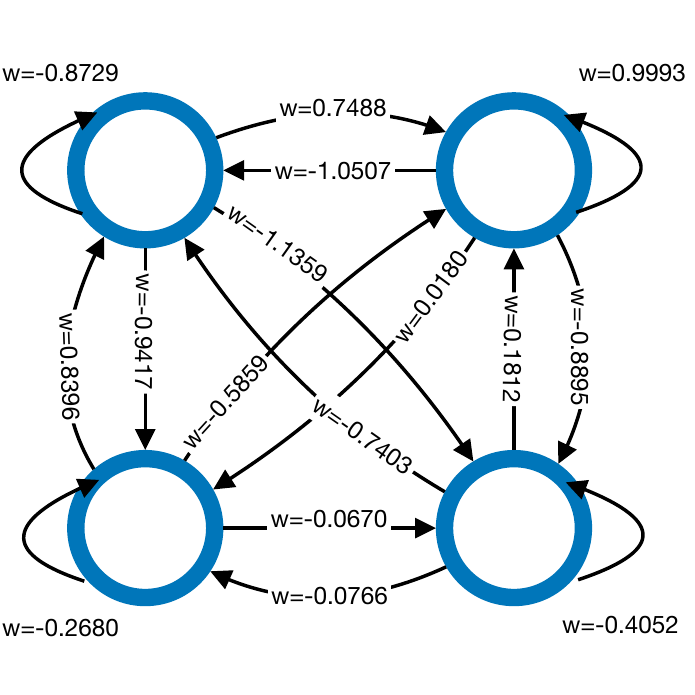}
    \end{minipage} \\
\midrule
5 & 0.739 & 0.628 & 92.078 &
$\mb{K} = \begin{bmatrix}
-0.8719 & 0.7481 & -1.1350 & -0.9407 & -1.0496\\
0.9987 & -0.8884 & 0.0227 & -0.7391 & 0.5061\\
-0.9542 & 0.0344 & 0.0141 & -0.1119 & -0.0735\\
-0.0193 & -0.5966 & -1.0733 & -0.0194 & 0.5313\\
-0.5180 & 0.5525 & 0.0412 & 0.4469 & 0.9878\\
\end{bmatrix}$ 
&\begin{minipage}{.18\textwidth}
      \includegraphics[width=\linewidth]{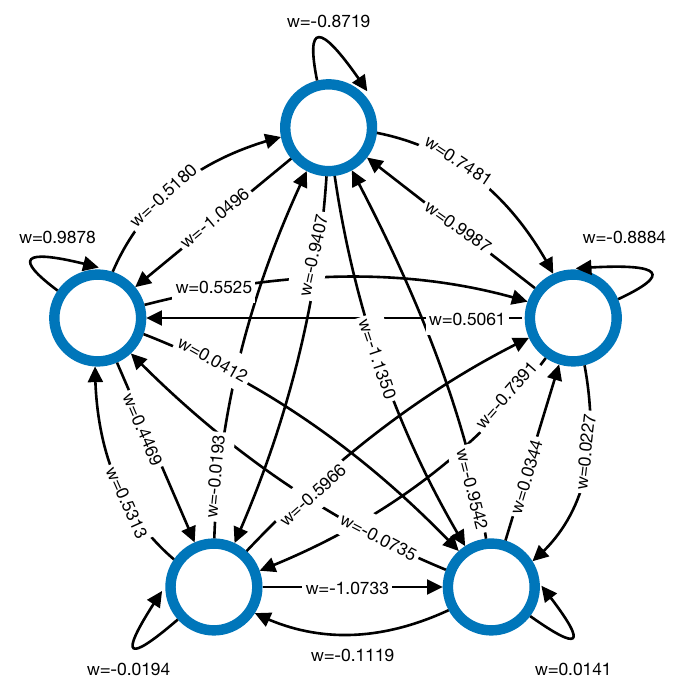}
    \end{minipage} \\
\bottomrule
\end{tabular}
}
\vspace{-5pt}
\end{table*}

\subsection{Experimental Results and Ablation Studies}

The main experimental results obtained by the comparison methods on the four brain graph datasets are reported in Table~\ref{tab:main_result}, where the scores are the averaged accuracy and F1 scores obtained via cross validation. On each dataset, we highlight the best method achieving the highest accuracy/F1 score with the bolded font. Also for the top-3 methods, we also indicate their relative rankings with the blue-colored numbers in parentheses attached next to the scores.

\subsubsection{Main Experimental Results} 
According to the results in the table, {\our} out-performs the other comparison methods on majority of the datasets with significant improvements. For instance, on the HIV-fMRI dataset, the average accuracy score obtained by {\our} is $0.805$. The score is about $50\%$ higher than the score obtained by classic graph learning methods {\wl} and {\freq}; $35\%$ higher than {\cnn} and {\gcn}. Compared with the related methods {\isonn} and {\capsule}, the accuracy score obtained by {\our} also has a $5\%$ and $8\%$ improvements, respectively. Similar performance improvements have also been observed with the F1 metric on the HIV-fMRI dataset.

For some of the datasets, the accuracy and F1 score obtained by {\our} can also be slightly lower than the other comparison methods. For instance, on ADHD-fMRI, the accuracy of {\our} is ranked at the 3rd position, which is lower than {\isonn} and {\fastisonn}. On HIV-DTI dataset, the F1 score of {\our} is also at the 3rd position, lower than {\isonn} and {\freq}. Partial reasons of such degradation can be caused by the reconstruction loss incorporated in {\our} for the capsule layers. According to our experimental studies, although removing the reconstruction loss will lead to high scores for {\our} for the above two cases and make {\our} increase to the best method as well, but it will also render the performance of {\our} to be unstable and will also suffer from overfitting on other datasets. Therefore, we preserve the graph reconstruction components as a necessary task in the {\our} model learning loss function by default for all the remaining experiments.

\subsubsection{Ablation Studies} 

The comparison of {\our} with {\isonn} and {\capsule} will also serve as the ablation studies of {\our}, {\isonn} discards the capsule layers from {\our} and {\capsule} can also be viewed as a {\our} variant without graph isomorphic feature extraction layers. Comparison between {\our} and {\isonn} indicates that, besides the sub-graph pattern existence scores, sub-graph orientation also plays an important role in learning graph representations. Incorporating the orientation into the model representation learning will improve the learned graph representations and lead to better classification results. Meanwhile, comparison between {\our} and {\capsule} indicates that the isomorphic features learned in {\our} will be better than the {\cnn} based feature extraction component in graph learning problems. The complementary components from {\isonn} and {\capsule} both contribute to the outstanding performance of {\our}.

\subsubsection{Other Interesting Observations}

Besides the comparison between {\our} with other methods, we also have some interesting observations by comparing other baseline methods with each other on the graph classification task. For instance, by incorporating the node re-ordering mechanisms, still based on the {\cnn} architecture, {\sdbn} and {\dgcnn} will obtain better performance than the vanilla {\cnn}. Without considering the node order issues, {\capsule} also out-performs {\cnn} with the capsule layers that capture the orientations of the learned kernels, especially on the HIV-fMRI and HIV-DTI datasets. Also the recent graph neural network based models {\gcn}, {\gin} and {\diffpool} also outperforms the classic graph learning models {\wl} and {\freq} with large improvements. It demonstrates the automatic graph representation learning will lead to better performance than the classic manual feature extraction in graph learning.

\begin{figure*}[t]
    \begin{minipage}{\textwidth}

    \centering
    	\includegraphics[width=0.8\linewidth]{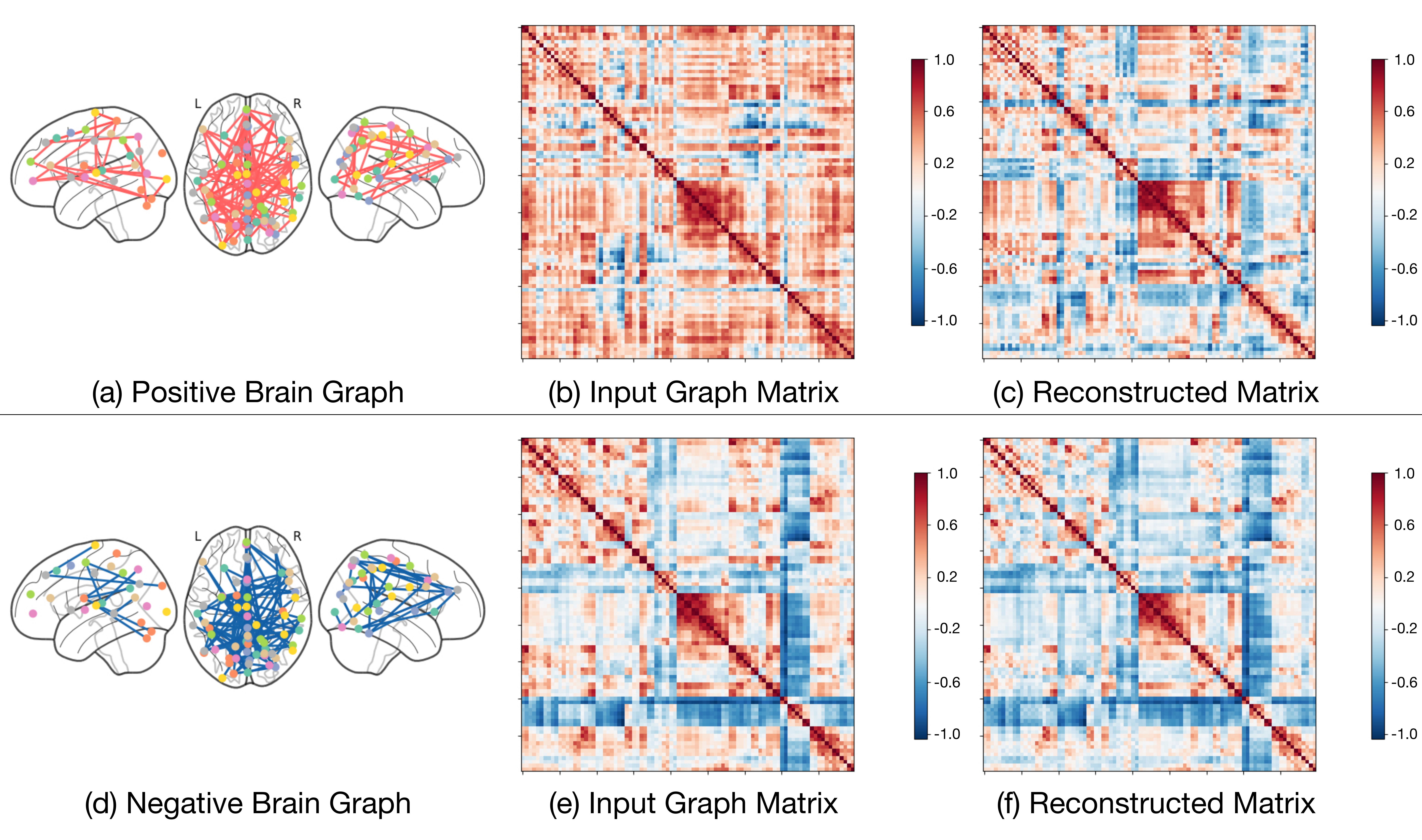}

    	\caption{Case Studies on Brain Graph Reconstruction in {\our} on HIV-fMRI Dataset. Plots (a): an input positive brain graph from the HIV-fMRI dataset; (b): input positive brain graph adjacency matrix; (c): adjacency matrix reconstructed by {\our} of the positive brain graph. Plots (d): an input negative brain graph from the HIV-fMRI dataset; (e): input negative brain graph adjacency matrix; (f): adjacency matrix reconstructed by {\our} of the negative brain graph.}
    	\label{fig:case_study}
    \end{minipage}%

\end{figure*}

\subsection{Experimental Analysis and Studies}

Besides the main results and ablation studies discussed above, we will also provide some more detailed analyses about the {\our} model proposed in this paper in terms of the hyper-parameter $k$ (sub-graph template size) and the brain graph reconstruction with the learned class-wise digit capsule. 

\subsubsection{Kernel Size Analysis}

As illustrated in Table~\ref{tab:parameter_analysis}, we study the impacts of the hyper-parameter $k$ on the {\our} model learning performance. Specifically, we change the parameter $k$ with values from $\{1, 2, 3, 4, 5\}$ and analyze the learning results of {\our} in terms of the accuracy and F1 based score evaluations, time costs, and the learned sub-graph template by the {\our}.

\noindent \textbf{Learning Performance}: According to the results in Table~\ref{tab:parameter_analysis}, as parameter $k$ increases from $1$ to $5$, the accuracy and F1 score obtained by {\our} will increase first and then decrease, which can achieve the best performance at $k=4$. To be more specific, when $k$ is small, e.g., $k=1$ or $k=2$, the sub-graph template we can learn for the isomorphic feature extraction will be very simple with one or two nodes only, representations learned with which will be useless for the classification task. On the other hand, as $k$ increase to a large value, e.g., $k=5$, the sub-graph template we can learn in {\our} will be much more complicated. It is relatively difficult to learn a larger and complicated sub-graph template for the feature extraction. Also it is highly likely that we cannot match such complicated sub-graph templates with any regions of the input brain graphs, which renders the features learned by the isomorphic layer in {\our} to be less useful. It explains why a larger $k$ will degrade the learning performance as well.

\noindent \textbf{Time Consumption}: The parameter $k$ decides the size of sub-graph templates to be learned. As $k$ increases, the learning time cost will increase steadily. The main reason is, for a larger parameter $k$, the number of permutation matrix $P$, i.e., $k!$, to be enumerated in the {\our} model will grow exponentially, which leads to more training costs. For instance, when $k=4$, the learning time consumption of {\our} is only $10.995$s. Meanwhile, when $k$ increases to $5$, its time cost will grow to $92.078$ instead. 

\noindent \textbf{Learned Sub-graph Template}: In the last two columns of the table, we provide the learned sub-graph template matrix and sub-graph structure. When $k=1$, its corresponding template matrix contains only one single value, which denotes the weight of self-edge pointing to the single node in the sub-graph. For the remaining $k$ values, i.e., $k=2$ to $k=5$, its template matrix will be a square matrix, whose corresponding sub-graph will be a fully connected graph instead. Such sub-graphs illustrate the patterns that may exist in the input brain graphs, which can be used for the isomorphic feature extraction. According to the learned template matrix and the sub-graph structure, the weights of the graph edges can be both positive and negative values. Since the brain graph edge denotes the activeness correlation among brain regions, the learned sub-graph template matrix indicate that both the positive and negative correlations will contribute to the isomorphic feature learning.

\subsubsection{Case Studies on Graph Reconstruction}

Finally, in Figure~\ref{fig:case_study}, we provide the case studies about two input brain graphs from the HIV-fMRI dataset (one has the positive label and the other one has the negative label). We illustrate their connection structures, corresponding input graph adjacency matrix and their reconstructed adjacency matrix by the {\our} model.

\noindent \textbf{Input Brain Graphs}: Formally, the input brain graphs have $90$ nodes, denoting the $90$ areas in the brain. The weighted connections both within and across the left and right cerebral hemispheres are illustrated by the edges in the graph. We use Nilearn\footnote{https://nilearn.github.io/stable/index.html} as the visualization toolkit, which will discard the edges whose weight absolute value is less than $0.05$. The edge colors in the brain graphs don't indicate the edge weight value or polarity, which are only used to differential the positive and negative graph instances. According to both the brain graphs, as well as the corresponding adjacency matrices, we observe that the positive and negative graph instances have very different graph structures and connection weights.

\noindent \textbf{Brain Graph Reconstruction}: In the right plots, we also illustrate the adjacency matrices reconstructed by {\our} from the class-wise digit capsules. The reconstruction loss term is added into the classification objective function for model learning regularization purposes. According to the matrix entries, although {\our} cannot exactly recovery the input matrix, but their differences compared with the input brain graph is still acceptable. The results illustrate that the class-wise digit capsule learned in {\our} carries information about the input brain graph, which can also help reconstruct the input brain graphs as well.

\section{Conclusion}\label{sec:conclusion}

In this paper, we have introduced a new representation learning model for brain graph data, namely {\our} (isomorphic capsule network). {\our} extends {\isonn} to extract not only the sub-graph existence confidence scores but also sub-graph pose and orientation basic properties in the learned representations. Via the dynamic routing and graph capsule learning, {\our} can effectively learn the digit capsules of the input brain graph instances. In addition to minimizing the differences between the floated label vector versus the ground truth, {\our} also involves the input brain graph reconstruction as a regularization task. Extensive experiments done on real-world brain graph benchmark datasets also demonstrate the effectiveness of the representations learned {\our} in classifying the brain graph instances.

\bibliography{reference}

\begin{thebibliography}{10}

\bibitem{Capsule_Tumor_Classification}
Parnian Afshar, Konstantinos~N. Plataniotis, and Arash Mohammadi.
\newblock Capsule networks for brain tumor classification based on {MRI} images
  and course tumor boundaries.
\newblock {\em CoRR}, abs/1811.00597, 2018.

\bibitem{AT16}
James Atwood and Don Towsley.
\newblock Search-convolutional neural networks.
\newblock {\em CoRR}, abs/1511.02136, 2015.

\bibitem{Bassett_Human_09}
Danielle Bassett and Edward Bullmore.
\newblock Human brain networks in health and disease.
\newblock {\em Current opinion in neurology}, 22:340--7, 07 2009.

\bibitem{Cao_Tensor_14}
Bokai Cao, Lifang He, Xiangnan Kong, Philip~S. Yu, Zhifeng Hao, and Ann~B.
  Ragin.
\newblock Tensor-based multi-view feature selection with applications to brain
  diseases.
\newblock In {\em Proceedings of the 2014 IEEE International Conference on Data
  Mining}, ICDM ?14, page 40?49, USA, 2014. IEEE Computer Society.

\bibitem{identifying_brain}
Bokai Cao, Xiangnan Kong, Jingyuan Zhang, Philip~S. Yu, and Ann~B. Ragin.
\newblock Identifying hiv-induced subgraph patterns in brain networks with side
  information.
\newblock {\em Brain Informatics}, 2(4):211--223, 2015.

\bibitem{Fallani_Graph_14}
Fabrizio De~Vico~Fallani, Jonas Richiardi, Mario Chavez, and Sophie Achard.
\newblock Graph analysis of functional brain networks: Practical issues in
  translational neuroscience.
\newblock {\em Philosophical transactions of the Royal Society of London.
  Series B, Biological sciences}, 369, 06 2014.

\bibitem{Gao_GraphNAS_19}
Yang Gao, Hong Yang, Peng Zhang, Chuan Zhou, and Yue Hu.
\newblock Graphnas: Graph neural architecture search with reinforcement
  learning.
\newblock {\em CoRR}, abs/1904.09981, 2019.

\bibitem{NIPS2019_8982}
Taeyoung Hahn, Myeongjang Pyeon, and Gunhee Kim.
\newblock Self-routing capsule networks.
\newblock In {\em Advances in Neural Information Processing Systems 32}, pages
  7656--7665. Curran Associates, Inc., 2019.

\bibitem{e2018matrix}
Geoffrey~E Hinton, Sara Sabour, and Nicholas Frosst.
\newblock Matrix capsules with {EM} routing.
\newblock In {\em International Conference on Learning Representations}, 2018.

\bibitem{Kipf_Semi_CORR_16}
Thomas~N. Kipf and Max Welling.
\newblock Semi-supervised classification with graph convolutional networks.
\newblock {\em CoRR}, abs/1609.02907, 2016.

\bibitem{Klicpera_Personalized_18}
Johannes Klicpera, Aleksandar Bojchevski, and Stephan G{\"{u}}nnemann.
\newblock Personalized embedding propagation: Combining neural networks on
  graphs with personalized pagerank.
\newblock {\em CoRR}, abs/1810.05997, 2018.

\bibitem{Kong_Brain_14}
Xiangnan Kong and Philip~S. Yu.
\newblock Brain network analysis: A data mining perspective.
\newblock {\em SIGKDD Explor. Newsl.}, 15(2):30?38, June 2014.

\bibitem{lalonde2018capsules}
Rodney LaLonde and Ulas Bagci.
\newblock Capsules for object segmentation.
\newblock {\em arXiv preprint arXiv:1804.04241}, 2018.

\bibitem{10.5555/646469.691875}
Yann LeCun, Patrick Haffner, L\'{e}on Bottou, and Yoshua Bengio.
\newblock Object recognition with gradient-based learning.
\newblock In {\em Shape, Contour and Grouping in Computer Vision}, page 319,
  Berlin, Heidelberg, 1999. Springer-Verlag.

\bibitem{Lee_Identifying_17}
John Lee, Xiangnan Kong, Yihan Bao, and Constance Moore.
\newblock Identifying deep contrasting networks from time series data:
  Application to brain network analysis.
\newblock In {\em Proceedings of the 2020 SIAM International Conference on Data
  Mining}, pages 543--551, 06 2017.

\bibitem{Lee2017IdentifyingDC}
John~Boaz Lee, Xiangnan Kong, Yi-Ting Bao, and Constance~M. Moore.
\newblock Identifying deep contrasting networks from time series data:
  Application to brain network analysis.
\newblock In {\em SDM}, 2017.

\bibitem{Lee_Deep_20}
John~Boaz Lee, Xiangnan Kong, Constance~M. Moore, and Nesreen~K. Ahmed.
\newblock Deep parametric model for discovering group-cohesive functional brain
  regions.
\newblock In {\em Proceedings of the 2020 SIAM International Conference on Data
  Mining}, 2020.

\bibitem{Li_Combinatorial_18}
Zhuwen Li, Qifeng Chen, and Vladlen Koltun.
\newblock Combinatorial optimization with graph convolutional networks and
  guided tree search.
\newblock {\em CoRR}, abs/1810.10659, 2018.

\bibitem{doi:10.1137/1.9781611974973.21}
Xinyue Liu, Xiangnan Kong, and Ann~B. Ragin.
\newblock {\em Unified and Contrasting Graphical Lasso for Brain Network
  Discovery}, pages 180--188.

\bibitem{MBBV15}
Jonathan Masci, Davide Boscaini, Michael~M. Bronstein, and Pierre
  Vandergheynst.
\newblock Shapenet: Convolutional neural networks on non-euclidean manifolds.
\newblock {\em CoRR}, abs/1501.06297, 2015.

\bibitem{Meng_Isomorphic_NIPS_19}
Lin Meng and Jiawei Zhang.
\newblock Isonn: Isomorphic neural network for graph representation learning
  and classification.
\newblock {\em CoRR}, abs/1907.09495, 2019.

\bibitem{Menon_Large_11}
Vinod Menon.
\newblock Large-scale brain networks and psychopathology: A unifying triple
  network model.
\newblock {\em Trends in cognitive sciences}, 15:483--506, 09 2011.

\bibitem{Mitra_Analysis_99}
P.P. Mitra and B.~Pesaran.
\newblock Analysis of dynamic brain imaging data.
\newblock {\em Biophysical Journal}, 76(2):691 -- 708, 1999.

\bibitem{MBMRSB16}
Federico Monti, Davide Boscaini, Jonathan Masci, Emanuele Rodol{\`{a}}, Jan
  Svoboda, and Michael~M. Bronstein.
\newblock Geometric deep learning on graphs and manifolds using mixture model
  cnns.
\newblock {\em CoRR}, abs/1611.08402, 2016.

\bibitem{NAK16}
Mathias Niepert, Mohamed Ahmed, and Konstantin Kutzkov.
\newblock Learning convolutional neural networks for graphs.
\newblock {\em CoRR}, abs/1605.05273, 2016.

\bibitem{Deep_Capsule_Net}
Jathushan Rajasegaran, Vinoj Jayasundara, Sandaru Jayasekara, Hirunima
  Jayasekara, Suranga Seneviratne, and Ranga Rodrigo.
\newblock Deepcaps: Going deeper with capsule networks.
\newblock {\em CoRR}, abs/1904.09546, 2019.

\bibitem{NIPS2017_6975}
Sara Sabour, Nicholas Frosst, and Geoffrey~E Hinton.
\newblock Dynamic routing between capsules.
\newblock In I.~Guyon, U.~V. Luxburg, S.~Bengio, H.~Wallach, R.~Fergus,
  S.~Vishwanathan, and R.~Garnett, editors, {\em Advances in Neural Information
  Processing Systems 30}, pages 3856--3866. Curran Associates, Inc., 2017.

\bibitem{Sabour_Capsule_17}
Sara Sabour, Nicholas Frosst, and Geoffrey~E. Hinton.
\newblock Dynamic routing between capsules.
\newblock {\em CoRR}, abs/1710.09829, 2017.

\bibitem{SGTHM09}
Franco {Scarselli}, Marco {Gori}, Ah~Chung {Tsoi}, Markus {Hagenbuchner}, and
  Gabriele {Monfardini}.
\newblock The graph neural network model.
\newblock {\em IEEE Transactions on Neural Networks}, 20(1):61--80, Jan 2009.

\bibitem{NIPS2009_0a49e3c3}
Nino Shervashidze and Karsten Borgwardt.
\newblock Fast subtree kernels on graphs.
\newblock In Y.~Bengio, D.~Schuurmans, J.~Lafferty, C.~Williams, and
  A.~Culotta, editors, {\em Advances in Neural Information Processing Systems},
  volume~22. Curran Associates, Inc., 2009.

\bibitem{singh2019dual}
Maneet Singh, Shruti Nagpal, Richa Singh, and Mayank Vatsa.
\newblock Dual directed capsule network for very low resolution image
  recognition.
\newblock In {\em Proceedings of the IEEE International Conference on Computer
  Vision}, pages 340--349, 2019.

\bibitem{Stam_Graph_08}
C.J. Stam, W~Haan, Andreas Daffertshofer, B.F. Jones, I~Manshanden, Anne-Marie
  van Cappellen~van Walsum, Teresa Montez, Jeroen Verbunt, Jan De~Munck, Bob
  Dijk, Henk Berendse, and Ph~Scheltens.
\newblock Graph theoretical analysis of magnetoencephalographic functional
  connectivity in alzheimer's disease.
\newblock {\em Brain : a journal of neurology}, 132:213--24, 11 2008.

\bibitem{Velickovic_Graph_ICLR_18}
Petar Veli{\v{c}}kovi{\'{c}}, Guillem Cucurull, Arantxa Casanova, Adriana
  Romero, Pietro Li{\`{o}}, and Yoshua Bengio.
\newblock {Graph Attention Networks}.
\newblock {\em International Conference on Learning Representations}, 2018.

\bibitem{10.5555/1756006.1953039}
Pascal Vincent, Hugo Larochelle, Isabelle Lajoie, Yoshua Bengio, and
  Pierre-Antoine Manzagol.
\newblock Stacked denoising autoencoders: Learning useful representations in a
  deep network with a local denoising criterion.
\newblock {\em J. Mach. Learn. Res.}, 11:3371–3408, dec 2010.

\bibitem{Wang_Structural_17}
Shen Wang, Lifang He, Bokai Cao, Chun-Ta Lu, Philip~S. Yu, and Ann~B. Ragin.
\newblock Structural deep brain network mining.
\newblock In {\em Proceedings of the 23rd ACM SIGKDD International Conference
  on Knowledge Discovery and Data Mining}, KDD ?17, page 475?484, New York, NY,
  USA, 2017. Association for Computing Machinery.

\bibitem{sdbn}
Shen Wang, Lifang He, Bokai Cao, Chun-Ta Lu, Philip~S. Yu, and Ann~B. Ragin.
\newblock Structural deep brain network mining.
\newblock In {\em Proceedings of the 23rd ACM SIGKDD International Conference
  on Knowledge Discovery and Data Mining}, KDD '17, page 475–484, New York,
  NY, USA, 2017. Association for Computing Machinery.

\bibitem{zero_shot}
Congying Xia, Chenwei Zhang, Xiaohui Yan, Yi~Chang, and Philip~S. Yu.
\newblock Zero-shot user intent detection via capsule neural networks.
\newblock {\em CoRR}, abs/1809.00385, 2018.

\bibitem{gin}
Keyulu Xu, Weihua Hu, Jure Leskovec, and Stefanie Jegelka.
\newblock How powerful are graph neural networks?
\newblock {\em CoRR}, abs/1810.00826, 2018.

\bibitem{Yan_GSpan_02}
Xifeng Yan and Jiawei Han.
\newblock Gspan: Graph-based substructure pattern mining.
\newblock In {\em Proceedings of the 2002 IEEE International Conference on Data
  Mining}, ICDM ?02, page 721, USA, 2002. IEEE Computer Society.

\bibitem{1184038}
Xifeng Yan and Jiawei Han.
\newblock gspan: graph-based substructure pattern mining.
\newblock In {\em 2002 IEEE International Conference on Data Mining, 2002.
  Proceedings.}, pages 721--724, 2002.

\bibitem{Ying2018HierarchicalGR}
Rex Ying, Jiaxuan You, Christopher Morris, Xiang Ren, William~L. Hamilton, and
  Jure Leskovec.
\newblock Hierarchical graph representation learning with differentiable
  pooling.
\newblock {\em ArXiv}, abs/1806.08804, 2018.

\bibitem{loopynet}
Jiawei Zhang.
\newblock Deep loopy neural network model for graph structured data
  representation learning.
\newblock {\em CoRR}, abs/1805.07504, 2018.

\bibitem{Zhang_GResNet_19}
Jiawei Zhang and Lin Meng.
\newblock Gresnet: Graph residual network for reviving deep gnns from suspended
  animation.
\newblock {\em CoRR}, abs/1909.05729, 2019.

\bibitem{Zhang2018AnED}
Muhan Zhang, Zhicheng Cui, Marion Neumann, and Yixin Chen.
\newblock An end-to-end deep learning architecture for graph classification.
\newblock In {\em AAAI}, 2018.

\bibitem{multi_label_capsule}
Xinsong Zhang, Pengshuai Li, Weijia Jia, and Hai Zhao.
\newblock Multi-labeled relation extraction with attentive capsule network.
\newblock {\em CoRR}, abs/1811.04354, 2018.

\bibitem{Towards_Capsule}
Wei Zhao, Haiyun Peng, Steffen Eger, Erik Cambria, and Min Yang.
\newblock Towards scalable and reliable capsule networks for challenging {NLP}
  applications.
\newblock {\em CoRR}, abs/1906.02829, 2019.

\bibitem{ZCZYLS18}
Jie Zhou, Ganqu Cui, Zhengyan Zhang, Cheng Yang, Zhiyuan Liu, and Maosong Sun.
\newblock Graph neural networks: {A} review of methods and applications.
\newblock {\em CoRR}, abs/1812.08434, 2018.

\end{thebibliography}
\bibliographystyle{plain}


\end{document}